\documentclass[11pt,a4paper]{article}
\usepackage[hyperref]{acl2020}
\usepackage{times}
\usepackage{latexsym}

\usepackage{microtype}

\usepackage{graphicx}
\graphicspath{{figures/}}
\usepackage{multirow}
\usepackage{amsmath}
\usepackage{amssymb}

\aclfinalcopy % Uncomment this line for the final submission
\def\aclpaperid{2139} %nter the acl Paper ID here

\title{Learning Implicit Text Generation via Feature Matching}
\author{Inkit Padhi, Pierre Dognin, Ke Bai$^{\ddagger}$, Cicero Nogueira dos Santos$^{\dagger}$\thanks{*work done prior to joining Amazon}\\ \textbf{Vijil Chenthamarakshan, Youssef Mroueh, Payel Das}\\
IBM Research, $^{\ddagger}$Duke University, $^{\dagger}$Amazon AWS AI\\
  \texttt{inkpad@ibm.com, ke.bai@duke.edu, cicnog@amazon.com}\\ 
 \texttt{\{pdognin,ecvijil,mroueh,daspa\}@us.ibm.com} \\
  \\}
\date{}

\begin{document}
\maketitle
\begin{abstract}
Generative feature matching network (GFMN) is an approach for training implicit generative models for images by performing moment matching on features from pre-trained neural networks. In this paper, we present new GFMN formulations that are effective for sequential data. Our experimental results show the effectiveness of the proposed method, SeqGFMN, for three distinct generation tasks in English:
unconditional text generation,
class-conditional text generation, 
and unsupervised text style transfer.
SeqGFMN is stable to train and outperforms various adversarial approaches for text generation and text style transfer.
\end{abstract}

\section{Introduction}

Generative feature matching networks (GFMNs)  \cite{santos2019generative} has been recently proposed for learning implicit generative models by performing moment matching on features from pre-trained neural networks. This approach demonstrated that GFMN could produce state-of-the-art image generators while avoiding instabilities associated with adversarial learning.
Similarly to training generative adversarial networks (GANs) \cite{goodfellow2014generative},
GFMN training requires to backpropagate through the generated data to update the model parameters.
This backpropagation through the generated data, combined with adversarial learning instabilities, has proven to be a compelling challenge when applying GANs for discrete data such as text.
However, it remains unknown if this is also an issue for feature matching networks since the effectiveness of GFMN for sequential discrete data has not yet been studied.

In this work, we investigate the effectiveness of GFMN for different text generation tasks. As a \textbf{first contribution}, we propose a new formulation of GFMN for unconditional sequence generation, which we name \emph{Sequence-GFMN} or  \emph{SeqGFMN} for short, by performing token level feature matching. SeqGFMN has a stable training because it does not concurrently train a discriminator, which in principle could easily learn to distinguish between one-hot and soft one-hot representations.
As a result, we can use soft one-hot representations that the generator outputs during training without using the Gumbel softmax or REINFORCE algorithm as needed in GANs for text. Additionally, different from GANs \cite{zhu2018texygen}, SeqGFMN can produce meaningful text without the need of pre-training the generator with maximum likelihood estimation (MLE). We perform experiments using Bidirectional Encoder Representations from Transformers (BERT), GloVe, and FastText as our feature extractor networks. We use two different corpora, and assess both the quality and diversity of the generated texts with three different quantitative metrics: BLEU, Self-BLEU and Fr\'echet Infersent Distance (FID). Additionally, we show that the \emph{latent space} induced by SeqGFMN contains semantic and syntactic structure, as evidenced by interpolations in the \emph{z} space.

Our \textbf{second contribution} consists in proposing a new strategy for class-conditional generation with GFMN.
The key idea here is to perform class-wise feature matching.
We apply SeqGFMN to perform sentiment-based conditional generation using the Yelp Reviews dataset, and assess its performance using classification accuracy, BLEU, and Self-BLEU.

Finally, as a \textbf{third contribution}, we demonstrate that the feature matching loss is an effective approach to perform distribution matching in the context of unsupervised text style transfer (UTST).
Most previous work on UTST adapts the autoencoder framework by adding an additional loss term: adversarial loss or back-translation loss.
Our method consists in replacing the adversarial and back-translation loss with style-wise feature matching.
Our experimental results indicate that the feature matching loss produces better results than the traditionally used losses.

\section{Feature Matching Nets for Text}
\begin{figure*}[!ht]
\centering
    \includegraphics[width=1\textwidth]{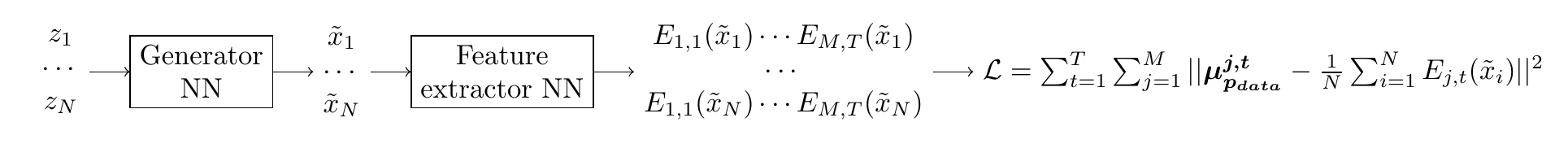}
   
    %<acl-submission>
    \caption{For each training iteration, Generator ($G$) outputs $N$ sentences from noise signals $z_1 \cdots z_N$. A fixed feature extractor is used to extract token level features ($E_{j,t}$) for the generated data. 
    %The feature matching loss is computed, and the error is backpropagated to parameters of the generator. 
    $\mathcal{L}$ is the $L_2$-norm of the difference between extracted features means of generated and real data $\boldsymbol{\mu_{p_{data}}^{j,t}}$, which is then backpropagted to update the parameters of $G$.
    %We precompute $\boldsymbol{\mu_{p_{data}}^{j,t}}$ on the entire real dataset (it does not change during training); the mean of generated data is estimated on a minibatch of size $N$. 
    The same strategy is used for variance terms in $\mathcal{L}$ (here ignored for brevity).
    }
    %</acl-submission>
    
    %\caption{For each training iteration, the Generator outputs $N$ sentences from noise signals $z_1 \cdots z_N$. A fixed feature extractor is used to extract token level features ($E_{j,t}$) for the generated data. Next, the token-level feature matching loss is computed, and the error is backpropagated to update the parameters of the Generator. $\mathcal{L}$ is the $L_2$-norm of the difference between extracted features means of generated and real data, $\boldsymbol{\mu_{p_{data}}^{j,t}}$. We precompute $\boldsymbol{\mu_{p_{data}}^{j,t}}$ on the entire real dataset (it does not change during training); the mean of generated data is estimated on a minibatch of size $N$. 
    %The same strategy is used for variance terms in $\mathcal{L}$.}
    
    \label{fig:gfmn}
\end{figure*}
\subsection{SeqGFMN}
Let $G$ be a sequence generator implemented as a neural network with parameters $\theta$, and let $E$ be a pretrained NLP feature extractor network with $L$ hidden layers, that produces features at token-level for each token in a sequence of length $T$.
The method consists of training $G$ by minimizing the following token-level feature matching loss function:
\begin{equation}
\min_{\theta}\sum_{t=1}^T \sum_{j=1}^M||\mu^{j,t}_{p_{data}} -\mu^{j,t}_{p_{G}}(\theta)||^2 + ||\sigma^{j,t}_{p_{data}} -\sigma^{j,t}_{p_{G}}(\theta) ||^2
\label{eq:lossG}
\end{equation}
where:
%\vskip -0.2in
%\begin{footnotesize}
\begin{align*}
\mu^{j,t}_{p_{data}}&=\mathbb{E}_{x\sim p_{data}}E_{j,t}(x) \in \mathbb{R}^{d_j}, \\
%\mu^j_{p_{data}}&=\underset{{x\sim p_{data}}}{\mathbb{E}}E_j(x) \in \mathbb{R}^{d_j} \\
\mu^{j,t}_{p_{G}}(\theta)&=\mathbb{E}_{z\sim \mathcal{N}(0,I_{n_z})}E_{j,t}(G(z;\theta)) \in \mathbb{R}^{d_j}, \\
\sigma^{j,t}_{p_{data},\ell}&=\mathbb{E}_{x\sim p_{data}}E_{j,\ell,t}(x)^2 - [\mu^{j,\ell,t}_{p_{data}}]^2,\\%\ell=1\dots d_j, \\
\sigma^{j,t}_{p_{G},\ell}(\theta)&=\mathbb{E}_{z\sim\mathcal{N}(0,I_{n_z})}E_{j,\ell,t}(G(z;\theta))^2 - [\mu^{j,\ell,t}_{p_{G}}]^2,\\
&\phantom{{}={}} \ell=1\dots d_j,
\end{align*}
%\end{footnotesize}
where $||.||^2$ is the $L_2$ loss;
$x$ is a real data point sampled from the data %generating 
distribution $p_{data}$;
$z \in \mathbb{R}^{n_z}$ is a noise vector sampled from the normal distribution $\mathcal{N}(0,I_{n_z})$;
$E_{j,t}(x)$ denotes the token-level $t$ feature map at a hidden layer $j$ from $E$;
$M\!\leq\!L$ is the number of hidden layers used to perform feature matching; $T$ is the maximum sequence length; and $\sigma^{2}_{p_{data}}$ and $\sigma^{2}_{p_{G}}$ are the variances of the features for real data and generated data respectively.
Note that this loss function is quite different from both the MLE loss used in regular language models and the adversarial loss used in GANs. 
%The main difference from Sequence GFMN with respect to GFMN  \cite{santos2019generative} is in incorporating this dense (per-token) feature matching at the token level. 

In order to train $G$, we first precompute $\mu^{j,t}_{p_{data}}$ and $\sigma^{j,t}_{p_{data},\ell}$ on the entire training data. During training, we generate a minibatch of \emph{fake} data by passing the Gaussian noise vector through the generator. 
The fixed feature extractor $E$ is used to extract features on the output of the generator at a per-token level. 
The loss is then computed, as mentioned in Eq.~\ref{eq:lossG}. The parameters $\theta$ of the generator G are optimized using stochastic gradient descent.
Note that the network $E$ is used for feature extraction only and is kept fixed during the training of $G$.
Similar to \cite{santos2019generative},
we use ADAM moving average, which allows us to use small minibatch sizes.
Fig.~\ref{fig:gfmn} illustrates SeqGFMN training; note that we use mean matching only for brevity, in practice we match both mean and diagonal covariance.
%We have illustrated mean matching in fig.~\ref{fig:gfmn} only for brevity; in practice, we match diagonal covariance as well. 

In our SeqGFMN framework,
the output of the generator $G$ is a sequence $\tilde{x}$ of \emph{soft one-hot representations}, 
$\{\tilde{w}_1, \tilde{w}_2, ..., \tilde{w}_T\}$,
where each element $\tilde{w}_i$ consists in the output of the softmax function at token $i$.
In the feature extractor $E$,
these soft one-hot representations are multiplied by an embedding matrix to generate \emph{soft embeddings}, which are then fed to the following layers of $E$.
\label{secMethods}

\subsection{Class-Conditional SeqGFMN}
\label{sec:cond_gen}
Conditional generation is motivated by the assumption that if the training data can be clustered into distinct and meaningful classes, knowledge of such classes at training time would improve the overall performance of the model. 
For class-based text generation, some datasets provide such opportunity by labeling the training data with relevant classes (e.g., positive/negative sentiment for Yelp Reviews dataset), information that can be leveraged by our model to condition the generation.

%For class-specific text generation, we can leverage the class information of the training data (e.g., positive/negative sentiment).
For this to be effective, the extracted features used for SeqGFMN need to be sufficiently representative of the text generated yet still be different between classes.
To account for the knowledge of latent classes, we extend the loss from Eq.\ref{eq:lossG} for the case of two distinct classes:
\begin{align}
\min_{\theta}\sum_{t=1}^T \sum_{j=1}^M & ||\delta^{j,t}_{c=0}||^2 + ||\Delta^{j,t}_{c=0} ||^2 +  \nonumber \\
& ||\delta^{j,t}_{c=1}||^2 + ||\Delta^{j,t}_{c=1} ||^2  
\label{eq:condlossG}
\end{align}

where $\delta^{j,t}_{c} = \mu^{j,t}_{p_{data}^c} -\mu^{j,t}_{p_{G}^c}(\theta)$ and 
$\Delta^{j,t}_c = \sigma^{j,t}_{p_{data}^c} -\sigma^{j,t}_{p_{G}^c}(\theta)$
follows the same definition for means and variances as Eq.\ref{eq:lossG}, with the exception that they are now class-dependent. 
%This extension implies that the loss takes into account the feature matching of both classes equally. This enables us to leverage the class information in the Generator.
Given a class $c$, we allow for conditional generation by conditioning the noise vector $z$ on $c$. Indeed, if $z\!\sim\!\mathcal{N}\left(0,I_{n_z}\right)$, applying a class dependent linear transformation $z_c=A_{c}z+b_{c}$ will change the noise distribution such that $z_c\sim\mathcal{N}\left(b_c,A_c^{\top}A_c\right)$. 
$A_c$ and $b_c$ are learned at training time so to minimize our loss. This enables the model to effectively sample a new input noise from distinct distributions, conditioned on the class $c$. Since the model can update the linear transformation parameters $A_c$ and $b_c$ to minimize its loss, the model can learn transformations that separate or disentangle between the different classes $c$ naturally. For example, conditioning on sentiment where $c\!=\!0$ is the negative sentiment class and $c\!=\!1$ the positive class, amounts simply to learning two transformations ($A_0$, $b_0$) and ($A_1$, $b_1$). 
%We can extend this approach to learn more than just linear transformations and allow for a deep neural network to be employed. 
This approach can be extended beyond learning linear transformations to allow for deep neural network to be employed. 
%For simplicity, we just address linear transformations in this paper.
During training, a minibatch is composed of input noise samples conditioned on class $c$. 
Within our generator, we use a conditional batch normalization (condBN) from \cite{cond_batch_norm}. The conditional BN is a 2-stage process: First, we perform a standard BN of a minibatch regardless of $c$ where $y_i = \text{BN}_{\gamma,\beta}(x_i)$, using notations from \cite{batch_norm}. Then $y_i$ enters a second stage where $w_i = \gamma_c y_i + \beta_c$ brings class dependency on $c$ as proposed in \cite{cond_batch_norm}. 
%Within the generator, we also use conditional batch normalization \cite{cond_batch_norm} which
This allows for the influence of class conditioning to carry over the whole model where conditional BN is used.  
%influences class conditioning over the whole model.
% where conditional BN is used. 
Our models can have three distinct configurations: conditional input noise, conditional BN, or both conditional input noise and conditional BN.

\subsection{Unsupervised Text Style Transfer (UTST) with SeqGFMN}
\label{sec:style_transfer}
% Text style transfer consists of rewriting a sentence from a given style $s_i$ into a different style $s_j$ while preserving the content and fluency.
% % and keeping the sentence fluent. 
% The major challenge for this task is the lack of parallel data, and many recent approaches adapt the encoder-decoder framework to work with non-parallel data \cite{shen_NIPS2017,fu:aaai18}.
% %This adaptation usually consists in using: (1) the reconstruction loss in an autoencoding fashion, which is intended to learn a conditional language model (decoder $D$) while providing content preservation; together with (2) a classification loss produced by a style classifier $C$, which is intended to guarantee the correct transfer. Balancing these two losses while generating good quality sentences is complicated, and several approaches such as adversarial discriminators \cite{shen_NIPS2017} and cycle-consistency loss \cite{melnyk2017} have been employed in recent works. Here, we use feature matching as a way to alleviate this problem.
% Essentially, our unsupervised text style transfer approach is an encoder-decoder trained with the following three losses:
Text style transfer consists of rewriting a sentence from a given style $s_i$ (e.g., informal) into a different style $s_j$ (e.g., formal) while maintaining the content and keeping the sentence fluent. 
The major challenge for this task is the lack of parallel data, and many recent approaches adapt the encoder-decoder framework to work with non-parallel data \cite{shen_NIPS2017,fu:aaai18}.
This adaptation normally consists in using: (1) the reconstruction loss in an autoencoding fashion, which is intended to learn a conditional language model (decoder $D$) while providing content preservation; together with (2) a classification loss produced by a style classifier $C$, which is intended to guarantee the correct transfer. 
Balancing these two losses while generating good quality sentences is difficult, and several approaches such as adversarial discriminators \cite{shen_NIPS2017} and cycle-consistency loss \cite{melnyk2017} have been employed in recent works. Here, we use feature matching as a way to alleviate this problem.
Essentially, our unsupervised text style transfer approach is an encoder-decoder trained with the following three losses:

\noindent \textbf{Reconstruction loss:} Given an input sentence $x^{s_i}$ from set $X$ and its decoded sentence $\hat{x}^{s_i} = D(E(x^{s_i}), s_i)$ (decoded in the same input style $s_i$), the reconstruction loss measures how well the decoder $D$ is able to reconstruct it:
\begin{align}
\label{eq:rec_loss}
\mathcal L_{\text{rec}} &= \mathbb{E}_{x^{s_i}\sim X} \left[-\log p_D(x^{s_i}| E(x^{s_i}), s_i)\right].
\end{align}

\noindent \textbf{Classification loss:} This loss is formulated as :
\begin{align}
\label{eq:class_td}
\mathcal L_{\text{class}} =& \mathbb{E}_{x^{s_i}\sim X} \left[-\log p_C(s_i|x^{s_i})\right] + \nonumber \\
& \mathbb{E}_{\hat{x}^{s_i \rightarrow s_j}\sim \hat{X}} \left[-\log p_C(s_j|\hat{x}^{s_i\rightarrow s_j})\right].
\end{align}
% where $\hat{X}$ is the set of style transferred sentences generated by the model. For the classifier, the first term provides supervised signal regarding style classification and the second term gives additional
% %training 
% signal from the transferred data, i.e generator's effectiveness on transferring texts to a different style. 
% %, enabling the classifier to be trained in a semi-supervised regime.
% %For the encoder-decoder the second term gives feedback on the current generator's effectiveness on transferring sentences to a different style.
where $\hat{X}$ is the set of style transferred sentences generated by the current model. For the classifier, the first term provides supervised signal regarding style classification and the second term gives additional training signal from the transferred data, enabling the classifier to be trained in a semi-supervised regime.
For the encoder-decoder the second term gives feedback on the current generator's effectiveness on transferring sentences to a different style.

\noindent \textbf{Feature Matching loss:} It is computed in a similar way as the class-conditional loss (Eq. \ref{eq:condlossG}).
This loss consists of matching statistics of the features for each style separately.
This means that
when transferring from style $s_i$ to $s_j$, 
we match the features of the resulting sentence with the features of real data that are from the target style $s_j$.

% \subsection{Feature Extractors for Textual Data}
% GFMN has demonstrated state-of-the-art results for image generation when deep CNNs pretrained on ImageNet \citep{russakovsky:2015} are used as feature extractors.
% The number and quality of features has a direct impact on the performance of GFMN, and the best results in \citet{santos2019generative} are obtained when features from all convolutional layers of both VGG and ResNet architectures are used for feature matching.
% Therefore, 
% we believe that an architecture that can produce a large number of high quality features would be beneficial for text generation: a deep neural network seems a natural fit for it.

% In this work,
% we experiment with different feature extractors that generate token-level representations.
% We use word embeddings from GloVe \cite{pennington-etal-2014-glove} and FastText \cite{bojanowski2017enrichingfasttext} as representatives of shallow (cheap-to-train) architectures.
% As a representative of large, deep feature extractor we use BERT \cite{devlin2018bert}.
% \citet{devlin2018bert} 
% demonstrated that the features extracted by BERT can boost the performance of diverse NLP tasks.
% Our hypothesis is that BERT features are informative enough to allow the training of (cross-domain) text generators with the help of feature matching.

\section{Related work}
\label{sec:relatedwork}

%\cite{zhu2018texygen} reviews and benchmarks several GAN models for text generation.
\cite{zhang17b_ftrMatchGanText} proposes
Adversarial Feature Matching for Text Generation by adding a reconstruction feature loss to the GAN objective. This is different from our setup, as our discriminator is not learned, and our feature matching is per token and not on a global sentence level. 
Sequence GAN (SeqGAN) \cite{yuZWY17_seqgan}, MaliGAN \cite{che17_maligan}, and RankGAN \cite{lin17_rankgan} use a pre-trained generator with MLE loss with a per token reward discriminator that is trained with reinforcement learning. 
%SeqGAN uses REINFORCE with many Monte-Carlo rollouts for training the generator. 
%MaliGAN normalizes the discriminator reward within a mini-batch. RankGAN replaces the binary cross-entropy loss of SeqGAN with a ranking loss. MaskGAN \cite{fedus2018maskgan} uses a recurrent discriminator and works on increasingly generated sequence sizes.  
SeqGFMN is similar to SeqGAN in the sense that it has a per token reward (per token feature matching loss). Still, it alleviates the need for pre-training the generator and the cumbersome training of a discriminator by relying on a fixed, state-of-the-art, text feature extractor such as BERT.
Due to the discrete nature of the problem, training implicit models is tricky \cite{de2019training}, which is addressed by using REINFORCE, actor-critic methods \cite{fedus2018maskgan}, and Gumbel softmax trick\cite{kusnerH16}.% for sequence generation is another alternative to allow backpropagation.
%We found that SeqGFMN trains properly with backpropagation using softmaxes with a temperature parameter. %More recently, RelGAN \cite{nie2019ICLR_relgan} showed that more complex architectures with relational and memory networks could improve the training of a text generator -- but still requires pre-training of the generator. 
%We did not find papers on class-conditional text generation using GANs.

For unsupervised text style transfer, different adaptations of the encoder-decoder framework have been proposed recently. \cite{shen_NIPS2017,fu:aaai18} uses adversarial classifiers to decode to a different style/language. \cite{melnyk2017},\cite{nogueira-dos-santos-etal-2018-fighting} proposed a method that combines a collaborative classifier with the back-transfer loss. \cite{prabhumoye2018style} presented an approach that trains different encoders, one per style, by combining the encoder of a pre-trained NMT and style classifiers. The main difference between our approach and these previous work consists in the fact that we use the feature matching loss to perform distribution matching.

\section{Experiments and Results}
\label{sec:experiments}
%\subsection{Experimental Setup}
%\vskip -0.07in
%\subsection{Datasets}
\noindent \textbf{Datasets}: We evaluate our proposed approach on three different english datasets: MSCOCO ~\cite{cocodataset}, EMNLP 2017 WMT News dataset~\cite{wmt17}, and Yelp Reviews Dataset \cite{shen_NIPS2017}. 
Both COCO and WMT News datasets are used for unconditional models, while Yelp Reviews is employed to evaluate class-conditional generation and unsupervised text style transfer. 
%We use 10K sentences for both training and testing, and for the COCO dataset, we use the same train/test split as in the Texygen framework~\cite{zhu2018texygen}. 
%We pad the sequences to have a fixed size. As we use a deconvolutional generator that produces the complete sentence at once, it learns to generate padded sentences. Padding is removed before computing the evaluation metrics.

%\subsection{Feature Extractors for Textual Data}
\noindent \textbf{Feature Extractors for Textual Data:}  
%GFMN has demonstrated state-of-the-art results for image generation when deep CNNs pre-trained on ImageNet \cite{russakovsky:2015} are used as feature extractors. The number and quality of features have a direct impact on the performance of GFMN, and the best results in \cite{santos2019generative} are obtained when features from all convolutional layers of both VGG and ResNet architectures are used for feature matching. Therefore, we believe that an architecture that can produce a large number of high-quality features would be beneficial for text generation: a deep neural network seems a natural fit for it.
We experiment with different feature extractors that generate token-level representations.
We use word embeddings from GloVe \cite{pennington-etal-2014-glove} and FastText \cite{bojanowski2017enrichingfasttext} as representatives of shallow (cheap-to-train) architectures.
As a representative of large, deep feature extractor we use BERT \cite{devlin2018bert}.
\citet{devlin2018bert} 
demonstrated that the features extracted by BERT can boost the performance of diverse NLP tasks.
Our hypothesis is that BERT features are informative enough to allow the training of (cross-domain) text generators with the help of feature matching.

%For unconditional and class-conditional experiments, we perform feature matching using features from all tokens of both generated texts and real texts. In the experiments on text style transfer, we perform feature matching using features from the \emph{[CLS]} token only. This is an artificial token used in the BERT model that, in principle, captures the representation of the entire sentence. For the sentiment transfer task, these features are enough to obtain good results. On the other hand, using features from \emph{[CLS]} token only does not perform well for unconditional and class-conditional generation.

%\subsection{Metrics}
\noindent \textbf{Metrics:}
In order to evaluate the diversity and quality of texts of the unconditional generators we use three metrics \emph{BLEU} \cite{Papineni:2002:BMA:1073083.1073135}, \emph{Self-BLEU}\cite{zhu2018texygen} and \emph{Fr\'echet Infersent Distance, FID}\cite{heuselRUNKH17FID}. Additionally, for class-conditional generation and unsupervised text style transfer, we report accuracy scores from a CNN sentiment classifier trained on the Yelp. 

\subsection{Experimental Results}

%\subsubsection{Unconditional Text Generation:}
\label{sec:unc_text_gen}

\begin{table*}[!ht]
\begin{center}
\begin{small}
\begin{tabular}{llcccccc}
%\hline
& Model & BLEU-2 & BLEU-3 & BLEU-4 & BLEU-5 & Self-BLEU & FID \\
%\hline
\hline
\multirow{9}{*}{COCO} & Real Data & 0.721 & 0.494 & 0.308 & 0.194 & 0.487 & 3.559 \\
\cline{2-8}
& SeqGAN & 0.044 & 0.019 & 0.012 & 0.010 & 0.026 & 13.167 \\
& MaliGAN &  0.042 & 0.017 & 0.011 & 0.008 & 0.032 & 15.855 \\
& RankGAN & 0.039 & 0.016 & 0.010 & 0.008 & \textbf{0.023} & 15.502 \\
& TextGAN &  0.034 & 0.015 & 0.010 & 0.008 & 0.624 & 17.275 \\
& RelGAN  & 0.230 & 0.055 & 0.026 & 0.017 & 0.811 & 13.948 \\
\cline{2-8}
& SeqGFMN (FastText)  & 0.389 & 0.153 & 0.089 & 0.059 & 0.644 & 6.371 \\
& SeqGFMN (Glove)    & 0.403 & 0.139 & 0.077 & 0.053 &  0.655 &  6.218 \\
& SeqGFMN (BERT)    & \textbf{0.695} & \textbf{0.476}  &  \textbf{0.277} & \textbf{0.186}  & 0.802 & \textbf{5.610} \\
\hline
%\hline
\multirow{9}{*}{WMT News} & Real Data & 0.852 & 0.596 & 0.356 & 0.199 & 0.289 & 0.365\\
\cline{2-8}
&  SeqGAN & 0.008 & 0.004 & 0.003 & 0.003 & 0.088 & 8.731 \\
& MaliGAN & 0.070 & 0.021 & 0.012 & 0.008 & \textbf{0.018} & 9.057 \\
& RankGAN & 0.188 & 0.055 & 0.024 & 0.015 & 0.973 & 12.306 \\
& TextGAN & 0.053 & 0.018 & 0.010 & 0.008 & 0.644 & 9.945 \\
& RelGAN  & 0.076 & 0.026 & 0.015 & 0.012 & 0.451 & 8.809 \\
\cline{2-8}
& SeqGFMN (FastText)  & 0.364 & 0.102 & 0.045 & 0.028 & 0.787 & 3.761 \\
& SeqGFMN (Glove) & 0.385 & 0.106 & 0.047 & 0.029 & 0.735 & 4.033 \\
& SeqGFMN (BERT) & \textbf{0.760} & \textbf{0.464} & \textbf{0.204} & \textbf{0.096} & 0.888 & \textbf{3.530} \\
\hline
\end{tabular}
\end{small}
\end{center}
\caption{Quantitative results for different implicit generators trained from scratch.}
\label{tab:res_from_scratch}
\end{table*}

%In this section, 
\textbf{\textit{Unconditional Text Generation}}: %We train SeqGFMN using three different feature extractors: GloVe, FastText and BERT.
In Tab.~\ref{tab:res_from_scratch}, we show quantitative results for SeqGFMN trained on COCO and WMT News using different feature extractors.
As expected, BERT as a feature extractor gives better performance because of a more significant and richer features used.
%This is due to: (1) the more significant number of features per token extracted from BERT (9216 vs. 300 in GloVe/FastText); and (2) the self-attention mechanism in BERT, which results in rich features with information about the whole sequence, especially in the top layers. 
%In the next sections, we use BERT as the feature extractor for all the experiments with SeqGFMN.
%For GloVe and FastText, we use their default pre-trained word embeddings of size 300.

%In Tab.~\ref{tab:res_from_scratch}, we 
We also present a comparison with other implicit generative models for text generation from scratch. 
We compare SeqGFMN with five different GAN approaches:
SeqGAN \cite{yuZWY17_seqgan},
MaliGAN \cite{che17_maligan},
RankGAN \cite{lin17_rankgan},
TextGAN \cite{zhang17b_ftrMatchGanText}
and RelGAN \cite{nie2019ICLR_relgan}.
%We train all GAN models (except RelGAN) using Texygen\footnote{https://github.com/geek-ai/Texygen} tool \cite{zhu2018texygen}.
%For RelGAN, we train the model using the author's code base\footnote{https://github.com/weilinie/RelGAN}.
We do not use generator pre-training for any of the models.
%We used default parameters for all the models, with the exception that we skipped the pre-training step.
%In Tab.~\ref{tab:res_from_scratch}, we report BLEU, Self-BLEU, and FID results for SeqGFMN and the five GAN approaches trained on COCO and WMT News.
As reported in Tab.~\ref{tab:res_from_scratch}, SeqGFMN outperforms all GAN models in terms of BLEU and FID.
The combination of low BLEU and low Self-BLEU for the different GANs indicates that the learned models generate random n-grams that do not appear in the test set.
All GANs fail to learn reasonable models due to the challenges of learning a discrete data generator from scratch under the min-max game.
Whereas, SeqGFMN can learn suitable generators without the need of generator pre-training.

%\label{sec:expts_cond_gen}

\noindent \textbf{\textit{Class-conditional Generation}}: Conditional generation experiments were conducted on Yelp Reviews dataset with sentiment labels (178K negative, 268K positive). %for a 40\%-60\% negative/positive split
For this experiment, we first pre-trained the Generator using a conditional denoising AE where class labels are provided only to the decoder $D$. %(with word dropout probability of .25 in the input text). 
The architecture of the encoder is the same as in \cite{zhang17_deconv} with three strided convolutional layers. Once pre-trained, $D$ is used as initialization for our Generator $G$.
The training is similar to the previous section except now sentiment class labels are passed to $G$, and class-dependent statistics of BERT features are used, as described in \ref{sec:cond_gen}.
%During training, mini-batches are class-balanced, ensuring 50\%-50\% negative/positive labels. Sequence length is limited to 16 tokens. A regular, unconditional baseline model is built on the same data for reference. 

\begin{table}[ht]
\begin{center}
\begin{small}
\begin{tabular}{lcccc}
\hline \textbf{Model} & \textbf{Accu.} & \textbf{Class} & \textbf{BLEU3} & \textbf{Self-BLEU3} \\ 
\hline
Baseline &  -  & - & \textbf{0.415} & 0.509 \\
\hline
Conditional  & \textbf{0.746}    & 0 & \textbf{0.473} & 0.498 \\
Noise+BN              &          & 1 & \textbf{0.413} & 0.472 \\
\hline
Cond. BN       & 0.745    & 0 & 0.423 & 0.473 \\
               &          & 1 & 0.395 & 0.505 \\
\hline
Cond. Noise    & 0.495    & 0 & 0.413 & \textbf{0.458} \\
               &          & 1 & 0.412 & \textbf{0.470} \\
\hline
\end{tabular}
\end{small}
\end{center}
\caption{Comparison between Sentiment-dependent and class-agnostic (unconditional) SeqGFMN models. %Accuracy is measured from a pre-trained classifier.
}
\label{tab:sent_class}
\end{table}
Tab.~\ref{tab:sent_class} presents results for our regular model (baseline) and the three conditional generators: Cond. Noise, Cond. Batch Normalization (BN), Cond. Noise+BN. 
%From the BLEU-3 score of the regular model, it is clear that generating Yelp Reviews is a challenging task. 
We use 10K generated sentences for each sentiment class to compute classification accuracy. 
%The assumption is that texts capturing sentiment classes better will lead to better prediction accuracy from the classifier (i.e., classifier accuracy is somewhat correlated to conditional generation quality). 
%In terms of accuracy and BLEU-3 score, the Cond. Noise+BN model provides the best results, followed by Cond. BN and Cond. Noise. Clearly, the combination of conditional noise transformation and BN provides a generator better able to capture and leverage the sentiment class.  This model performs better than, or equivalently to, the unconditional model depending on the sentiment class.
In terms of accuracy and BLEU-3 score, the Cond. Noise+BN model provides the best generator as it is able to capture and leverage the class information.

% A cherry-picked interpolation between two input noises is presented in Tab.~\ref{tab:cond_interpolation}. 
% The sentences generated vary significantly in length, topic, and overall quality. Still, it remains within its sentiment class, indicating that the model did indeed learn to leverage the sentiment classes and separated them in the input noise manifold.

%\subsubsection{Unsupervised Text Style Transfer (UTST)}
%\label{sec:expts_style}

\noindent \textbf{\textit{Unsupervised Text Style Transfer (UTST)}}:
In Table \ref{tab:utst}, we report BLEU and accuracy scores for SeqGFMN and six baselines:
BackTranslation \cite{prabhumoye2018style}, which uses back-transfer loss;
CrossAligned \cite{shen_NIPS2017},
MultiDecoder \cite{fu:aaai18}, 
and StyleEmbedding \cite{fu:aaai18}, which use adversarial loss;
and TemplateBased \cite{li2018delete} and
Del-Retrieval \cite{li2018delete},
which uses rule-based methods.
%We selected these baselines because they do not perform pretraining of the decoder using pseudo parallel data, similar to our proposed method.
%We use the same train/validation/test partitioning from \cite{li2018delete}.
%Their test set consists of 1K sentences that were manually annotated, which means that the test set consists of parallel data. 
%Therefore, in Table \ref{tab:utst}, the BLEU score is computed between the transferred sentences and the human-annotated transferred sentences.
The BLEU score is computed between the transferred sentences and the human-annotated transferred references, similar to \cite{li2018delete}. And, the accuracy is based on our pre-trained classifier.
Compared to the other models, SeqGFMN produces the best balance between BLEU and accuracy.
Additionally, if we use back-transfer loss together with feature matching loss (\emph{SeqGFMN + BT}) our model gets a significant improvement on both metrics.

\begin{table}[ht]
\begin{center}
\begin{small}
\begin{tabular}{lcc}
\hline \textbf{Model} & \textbf{BLEU} & \textbf{Accuracy} \\ 
\hline
BackTranslation & 2.5 & 95.7 \\
CrossAligned & 9.1 & 74.1 \\
MultiDecoder & 14.6 & 50.1 \\
StyleEmbedding & 21.1 & 9.2 \\
TemplateBased & 22.6 & 81.1 \\
Del-Retrieval & 16.0 & 88.2 \\
\hline
SeqGFMN & 23.7 & 92.9 \\
SeqGFMN + BT & \textbf{24.5} & \textbf{96.4} \\
\hline
\end{tabular}
\end{small}
\end{center}
\caption{Comparison between SeqGFMN and other models for unsupervised text style transfer.}
\label{tab:utst}
\end{table}

% \begin{table*}[!ht]
% \begin{center}
% \begin{tabular}{l|c|cccccc}
% \hline
% & Ftr. Extractor & BLEU-2 & BLEU-3 & BLEU-4 & BLEU-5 & Self-BLEU3 & FID \\
% \hline
% \hline
% \multirow{3}{*}{COCO} & FastText  & 0.389 & 0.153 & 0.089 & 0.059 & \textbf{0.644} & 6.371 \\
% & Glove    & 0.403 & 0.139 & 0.077 & 0.053 &  0.655 &  6.218 \\
% & BERT   & \textbf{0.695} & \textbf{0.476}  &  \textbf{0.277} & \textbf{0.186}  & 0.802 & \textbf{5.610} \\
% \hline\hline
% \multirow{3}{*}{WMT News} & FastText  & 0.364 & 0.102 & 0.045 & 0.028 & 0.787 & 3.761 \\
% & Glove & 0.385 & 0.106 & 0.047 & 0.029 & \textbf{0.735} & 4.033 \\
% & BERT & \textbf{0.760} & \textbf{0.464} & \textbf{0.204} & \textbf{0.096} & 0.888 & \textbf{3.530} \\
% \hline
% \end{tabular}
% \end{center}
% \caption{Quantitative results for SeqGFMN generators that use different feature extractors.}
% \label{tab:res_ftr_exts}
% \end{table*}

\section{Conclusion}
\label{sec:conclusion}

We presented new implicit generative models based on feature matching loss that are suitable for unconditional and conditional text generation.
Our results demonstrated that backpropagating through discrete data is not an issue for the training via matching distributions at the token level. SeqGFMN can be trained from scratch without the need for RL or Gumbel Softmax.
%For unsupervised text generation,
%we presented effective SeqGFMN models trained from scratch and without the need for RL or Gumbel Softmax.
%Additionally, we demonstrated that conditional generation could be easily achieved by performing class-wise feature matching.
This approach has allowed us to create effective models for unconditional generation, class-conditional generation, and unsupervised text style transfer.
We believe this work opens a new competitive avenue in the area of implicit generative models for sequential data. 

%we reported experimental results that demonstrate the effectiveness of the feature matching network for two conditional generation tasks: class-conditional generation and unsupervised text style transfer.
%We believe that the performance of SeqGFMN can be further enhanced by increasing the mini-batch size and using larger architectures for the generator and feature extractor (for example, using BERT$_{\text{LARGE}}$ with 24 layers.)

\bibliography{anthology,acl2020}
\bibliographystyle{acl_natbib}
\clearpage
\appendix
\section*{{\Large Appendices}}
\appendix

\begin{table*}[!ht]
\begin{small}
\begin{center}
\begin{tabular}{ll}
\hline 
\textbf{Model} & \multicolumn{1}{c}{\textbf{COCO}} \\ 
\hline
\multirow{4}{*}{SeqGFMN} & a 747 aircraft plane flying on a runway . \\
& a kitchen with a kitchen sink and a microwave on the counters . \\
& a bike flag showcasing a person sitting near a street sign . \\
& a bathroom with a toilet on the counter . \\
\hline 
\multirow{4}{*}{RelGAN} & fry up on a nuts cargo black tonic rocks kept cruising basket adorable graveyard . \\
& border itl washer table a an green with bmw suit heater down . his pushed \\
& docked sofas wave messy nursing , triple black school a continue plane siking bbq pickup . \\
& quadruple several lots a loft buckets vines a bullhorn the appliances sidewalk sidewalk . uniforms
\\ 
\hline
\textbf{Model} & \multicolumn{1}{c}{\textbf{WMT News}} \\ 
\hline
\multirow{4}{*}{SeqGFMN} & the ban did nothing but say voters were illegally investing their time at college and to take on your calls at  \\
 & [CONT.] court , " ross . announced . \\
 & in addition , 32 typical economies in this period are reportedly pledged to have trillion pledged in another \\
 & [CONT.]  time , typically , tens to millions in million in feed . \\
\hline 
\multirow{3}{*}{RelGAN} & should should children about about about states . \\
& inquiry matthew his s a about am . . \\
& appeal only over a ve about found . \\
\hline 
\end{tabular}
\end{center}
\caption{Randomly sampled sentences from generators trained from scratch on COCO and WMT News datasets.}
\label{tab:samples_unc}
\end{small}

\end{table*}

\begin{table*}[!ht]
\begin{center}
\begin{footnotesize}
\begin{tabular}{ll}
%\hline \textbf{Yelp Reviews Dataset} \\ 
\hline \multicolumn{1}{l}{{\textbf{Positive Sentiment generated $z_1$}}}  & \multicolumn{1}{l}{\textbf{Negative Sentiment generated from $z_0$}}\\ 
\hline
    full of good food                                               & everything is bad food  \\
    love this place                                                 & avoid this place \\
    good job                                                        & horrible ! \\
    just perfect because my entire menu was fabulous                & completely upset with the salon \\
    everything is good !                                            & disgusting\\
    the service staff is extremely welcoming - and my mom loved it  & the salon itself is very poor , and my mom admitted it \\
\hline
\end{tabular}
\end{footnotesize}
\end{center}
\caption{Sentences generated using conditional SeqGFMN trained on Yelp Reviews dataset.}
\label{tab:gen_sent_cond}
\end{table*}

\begin{table*}[!ht]
\begin{center}
\begin{scriptsize}
\begin{tabular}{ll}
\hline \multicolumn{1}{l}{\textbf{Positive Sentiment (Original)}}  & \multicolumn{1}{l}{\textbf{Negative Sentiment (Transferred)}}\\ 
\hline
\multirow{2}{*}{place was clean and well kept , drinks were reasonably priced .} & place was dirty and drinks were expensive and watered down . (GT)\\
& place was dirty and horribly kept , drinks were horribly priced .	(SeqGFMN)\\
\multirow{2}{*}{food is very fresh and amazing !} & food was old and stale . (GT) \\
& food was ridiculous , too . (SeqGFMN)\\
\multirow{2}{*}{this place reminds me of home !}	& this place reminds me why i want to go home . (GT) \\
& this jerk reminds me of trash . (SeqGFMN) \\

\hline \multicolumn{1}{l}{\textbf{Negative Sentiment (Original)}}  & \multicolumn{1}{l}{\textbf{Positive Sentiment (Transferred)}}\\ 
\hline
\multirow{2}{*}{the decor was seriously lacking . } &
the decor was nice . (GT) \\
& the decor was superb . (SeqGFMN) \\
\multirow{2}{*}{now the food : not horrible , but below average .} & now the food : not bad , above average . (GT) \\
& now the food is fantastic ! (SeqGFMN) \\
\multirow{2}{*}{i wish i could give less than one star .} & i wish there were more stars to give . (GT) \\
& i love getting them ! (SeqGFMN) \\
\hline
\end{tabular}
\end{scriptsize}
\end{center}
\caption{Examples of sentiment transferred texts using SeqGFMN. \textbf{(GT)} = ground truth produced by a human.}
\label{tab:utst_samples}
\end{table*}

\begin{table*}[!ht]
\begin{small}
\begin{center}
\begin{tabular}{l}
\hline 
\multicolumn{1}{c}{\textbf{COCO}} \\ 
\hline
a group of people sleeps in the street \\
a group of people standing in the street \\
a toy of people warming a street sidewalk \\
an automobile car lies on an short parking road \\
an automobile car lies on an green parking road \\
an automobile car lies on an green bike field \\
the automobile car lies on an green parking field \\
the automobile car is on an green parking field \\
\hline
\multicolumn{1}{c}{\textbf{WMT News}} \\ 
\hline
``although that might do nothing -i admit it- and i've invested time time at work,'' i tend to say it doesn do nothing. \\
``although the odds do it -i get it- and ross hasn always conceded his chance at it,'' i tend to say our odds are there. \\
reportedly upon the call to court, i get it, while romney has promised that his ban did nothing but say voters had better announce... \\ %their call
reportedly upon the call at court and i get it, while voters didn \#\#rem realize the ban was there. \\
the said pledge would take on one another day, sexually claiming to top the worst in your period at the academy. \\
the us has to feed two-thirds in one month, typically in the best \#\#quest best \#\#gist at the in \& in millions in.\\
this will cover two-thirds billion trillion in this period, possibly two-thirds - 63 0 in one months. \\
in addition, regulators selected millions in one years, potentially billions in another decade, possibly the bottom-profile economies ... \\
\hline 
\end{tabular}
\end{center}
\caption{Interpolation in the latent space $z$ of SeqGFMN models trained on COCO Image Captions and WMT News.}
\label{tab:interp_sgfmn}
\end{small}
\end{table*}

\section{Experimental Setup}
\noindent \textbf{SeqGFMN Generator:}
We use a deconvolutional generator that extends the decoder architecture proposed in (Zhang et al., 2017).
%\cite{zhang17_deconv},
It consists of three strided deconvolutional layers followed by cosine similarity between the \emph{generated} token embeddings and an embedding matrix.
Our adaptations are as follows:
(1) we added two convolutional layers after the second deconvolution;
(2) we added a self-attention layer 
% \cite{zhan2018_self_att_gan} 
before the last deconvolutional layer;
(3) we added a convolutional layer after the last deconvolutional layer;
(4) after the final convolution, 
we multiply the resulting token embeddings by the embedding matrix and apply the softmax function to generate a probability distribution over the vocabulary.
We use the embedding matrix from BERT model 
and this matrix is not updated during the training of seqGFMN.
%These changes extend the capacity of the deconvolutional generator and lead to better results.
The number of convolutional filters used is 400 with kernel size of 5.
%OBS: above, we really meant filters and kernel size.

\noindent \textbf{SeqGFMN Training:}
SeqGFMNs are trained with an ADAM optimizer for which most hyper-parameters are kept fixed across datasets. We use $n_z\!=\!100$ and minibatch size of 128. 
We use learning rates of $10^{-4}$ and $10^{-3}$ for updating $G$, and ADAM Moving Averages (AMA), respectively.
The generator is trained for about 100K iterations. 
%, depending on the dataset.

\noindent \textbf{Feature Extractor Details:}
In the experiments with GloVe and FastText, 
we used their default 300 dimension vectors pre-trained on 6 billion tokens from Wikipedia 2014 \& Gigaword 5, and English Wikipedia, respectively.
In the experiments with BERT,
we use BERT$_{\text{BASE}}$ model, which contains 12 layers and produces 768 features per token per layer.
When using a maximum sequence of 32,
that leads to a total 294,912 features.
%BERT also provides a larger architecture, BERT$_{\text{LARGE}}$, with 24 layers, but using this model results in longer training times. For more details about BERT, we refer to \cite{devlin2018bert}.

\section{Unconditional Text Generation}
An interesting comparison would be between SeqGFMN and GANs that use BERT as a pre-trained discriminator.
However,
GANs fail to train when a very deep network is used as the discriminator %\cite{santos2019generative}.
Moreover,
SeqGFMN also outperforms GAN generators even when shallow word embeddings (Glove / FastText) are used to perform feature matching. 
Pretrained word embeddings are normally used in GANs for text. 

In Tab.~\ref{tab:samples_unc},
we present randomly selected samples that were generated by SeqGFMN and RelGAN.
These samples corroborate the quantitative results and show that SeqGFMN can generate good text when trained from scratch. At the same time, the state-of-the-art method RelGAN is unable to generate reasonable text without pretraining.
% Although SeqGFMN produces a high Self-BLEU for WMT News dataset because this model has not converged completely when we collected the result, 
% it has an FID that is much lower than all the other models.

\section{Class-Conditional Generation}

In Tab.~\ref{tab:gen_sent_cond}, we present cherry-picked examples of generated text.
Interestingly, since our input noise $z$ is transformed according to sentiment $c$, we implicitly have a pairing between $z_0$ and $z_1$. Text generated from $z_0$ and $z_1$ are related to the same $z$. The effect of this implicit pairing can be seen in the examples where sentences seem somehow related, but of the opposite sentiment.
Qualitatively, conditional SeqGFMN models can leverage class information to improve generation. 

In Table \ref{tab:utst_samples},
we present samples of original and sentiment transferred sentences.
For each original sentence, we show the reference transferred sentence from the test set (done by a human) and the sentence that was transferred by SeqGFMN.
Similar to other recently proposed UTST methods,
the most successful cases of sentiment transfer are the ones where the transfer can be done by removing and replacing a few words of the sentence.
In Table \ref{tab:utst_samples}, the last example of each block are cases where SeqGFMN does not do a good job when significant changes in the original sentence are required to perform a more fluent sentiment transfer.

\section{Unsupervised Text Style Transfer}
The baselines are calculated with the data collected by %\cite{luo2019dual}
(Luo et al., 2019)~\footnote{https://github.com/luofuli/DualRL/tree/master/outputs/yelp}
and using Unsupervised NMT methods %\cite{zhang2018style}.
(Zhang et al., 2018).

\section{Interpolation}
%We investigate whether the latent space induced by SeqGFMN encapsulates syntactic and semantic information.
We interpolate in the latent space of SeqGFMN $z$ and check whether the sentences generated by the interpolation are syntactically and/or semantically related.
In detail, we sample two vectors $z_0$ and $z_1$ from the prior distribution $p_z$ and build intermediate points $z_\lambda = \lambda z_1 + (1-\lambda)z_0$. In Tab.~\ref{tab:interp_sgfmn},
we show samples from two interpolations,
on models trained on COCO and WMT news dataset.
%and the other trained on the WMT News dataset.
In both these cases, we notice that there exists some syntactic and/or semantic relationship between the sentences along the interpolating path.
This is supporting evidence that the latent space induced by SeqGFMN is meaningful, and related sentences are close together in this latent space.

\end{document}